\documentclass{article}

 \usepackage[dblblindworkshop, final]{neurips_2025}
\workshoptitle{Data on the Brain \& Mind}

\usepackage{amsmath,amsfonts,bm}

\def\eqref#1{equation~\ref{#1}}
\def\1{\bm{1}}

\def\vtheta{{\bm{\theta}}}

\def\vb{{\bm{b}}}

\def\vh{{\bm{h}}}

\def\vw{{\bm{w}}}

\def\vy{{\bm{y}}}

\def\mW{{\bm{W}}}

\DeclareMathAlphabet{\mathsfit}{\encodingdefault}{\sfdefault}{m}{sl}
\SetMathAlphabet{\mathsfit}{bold}{\encodingdefault}{\sfdefault}{bx}{n}

\def\sR{{\mathbb{R}}}

\newcommand{\Var}{\mathrm{Var}}

\usepackage[utf8]{inputenc} % allow utf-8 input
\usepackage[T1]{fontenc}    % use 8-bit T1 fonts
\usepackage{hyperref}       % hyperlinks
\usepackage{url}            % simple URL typesetting
\usepackage{booktabs}       % professional-quality tables
\usepackage{amsfonts}       % blackboard math symbols
\usepackage{nicefrac}       % compact symbols for 1/2, etc.
\usepackage{microtype}      % microtypography
\usepackage{xcolor}         % colors

\usepackage{natbib}
\usepackage{times}
\usepackage{amsmath}
\usepackage{adjustbox}
\usepackage{amssymb}
\usepackage{amstext}
\usepackage{amsthm}
\usepackage{array}
\usepackage{tabularx}
\usepackage[table,dvipsnames]{xcolor}
\usepackage{makecell}
\usepackage{mathtools}
\usepackage{bbm}
\usepackage{latexsym}
\usepackage{color}
\usepackage{graphicx}
\usepackage{subcaption}
\usepackage{enumerate}
\usepackage{float}
\usepackage{caption}
\usepackage{comment}
\usepackage{multirow}
\usepackage{booktabs}
\usepackage{algorithm}
\usepackage[noend]{algpseudocode}
\usepackage{fullpage}
\usepackage{tikz}
\usepackage[justification=centering]{caption}

\usepackage{listings}
\newcommand\halfopen[2]{\ensuremath{[#1,#2)}}

\title{Benchmarking Probabilistic Time Series Forecasting Models on Neural Activity}

\author{
    \textbf{Ziyu Lu}$^{1}$,\hspace{1mm}
    \textbf{Anna J. Li}$^{2}$,\hspace{1mm} 
    \textbf{Alexander E. Ladd}$^{2}$,\hspace{1mm} \textbf{Pascha Matveev}$^{2}$,\hspace{1mm} \textbf{Aditya Deole}$^{2}$,\hspace{1mm} \\
    \textbf{Eric Shea-Brown}$^{1,3,\dagger}$,\hspace{1mm}
    \textbf{J. Nathan Kutz}$^{1,4,\dagger}$,\hspace{1mm} \textbf{Nicholas A. Steinmetz}$^{2,\dagger}$ \\
    \small $^1$Department of Applied Mathematics, University of Washington, Seattle, WA, USA. \\
    \small $^2$Department of Neurobiology and Biophysics, University of Washington, Seattle, WA, USA. \\
    \small $^3$Allen Institute for Brain Science, Seattle, WA, USA. \\
    \small $^4$Department of Electrical and Computer Engineering, University of Washington, Seattle, WA, USA. \\
    \small $^\dagger$ Authors jointly supervised. \\
    \small Correspondence to: \texttt{\{luziyu, etsb, kutz, nsteinme\} @ uw.edu} 
}

\begin{document}

\maketitle

\begin{abstract}
%!TEX root = ../main.tex

Neural activity forecasting is central to understanding neural systems and enabling closed-loop control. While deep learning has recently advanced the state-of-the-art in the time series forecasting literature, its application to neural activity forecasting remains limited.
To bridge this gap, we systematically evaluated eight probabilistic deep learning models, including two foundation models, that have demonstrated strong performance on general forecasting benchmarks. We compared them against four classical statistical models and two baseline methods on spontaneous neural activity recorded from mouse cortex via widefield imaging. Across prediction horizons, several deep learning models consistently outperformed classical approaches, with the best model producing informative forecasts up to 1.5 seconds into the future. Our findings point toward future control applications and open new avenues for probing the intrinsic temporal structure of neural activity. 
\end{abstract}

%!TEX root = ../main.tex

\section{Introduction}
Predicting the future state of a system is central to understanding its underlying dynamics. Beyond its value from the basic science perspective, time series forecasting is also critical to a wide range of real-world applications -- including weather prediction\cite{price2025probabilistic}, renewable energy planning\cite{wang2019review}, and financial market analysis\cite{kumbure2022machine} -- guiding decisions ranging from everyday choices to high-stakes operations. Among the most challenging and rewarding domains for forecasting is the brain -- one of the most complex and sophisticated systems known to science. Accurate forecasting of brain activity can not only offer insights into the underlying neural mechanisms but also enable transformative applications, including early intervention for neurological disorders\cite{mormann2007seizure}, the development of therapeutic brain stimulation paradigms\cite{de2025brain}, and the advancement of brain–machine interfaces\cite{lebedev2006brain}.

The field of time series forecasting has evolved dramatically over the past decade, driven by the rise of machine learning and deep neural networks\cite{lim2021time}. Several studies have assembled datasets across various domains including energy, sales, climate, and healthcare, and systematically assessed the performance of both modern deep learning-based models and classical statistical methods on them\cite{makridakis2023statistical, makridakis2020m4, makridakis2022m5, godahewa2021monash, qiu2024tfb, shao2024exploring, aksu2024gift}. However, with the exception of one recent study\cite{lueckmann2025zapbench} which focuses on brain activity of zebrafish, neural time series have been largely absent from established forecasting benchmarks, and their markedly different characteristics from the included series make it unclear whether prior conclusions would apply to them. For instance, neural recordings are often sampled or binned at seconds to milliseconds resolutions, whereas most benchmark datasets are sampled at much coarser intervals (hourly, daily, or monthly). Furthermore, while oscillations are an important feature of brain activity\cite{thut2012functional}, there is also a substantial amount of activity that does not exhibit persistent trends or periodicity/seasonality. In contrast, many real-world time series, such as climate records or energy consumption, are dominated by these patterns. These differences highlight the importance of systematically reassessing forecasting methods in the context of neural data. 

Among neural recordings from different species, mouse data bring many important opportunities for time series forecasting. Foremost among the reasons is their relevance for developing and testing future neural control systems: recent advances in optogenetics\cite{zatka2021sensory,matveev2024simultaneous,lakunina2025neuropixels} have enabled targeted manipulation of brain activity in mice, making accurate forecasts immediately applicable to the design of closed-loop control systems\cite{grosenick2015closed}. Moreover, modern recording technologies -- such as Neuropixels probes\cite{steinmetz2021neuropixels, melin2024large} for high-density, single-neuron resolution measurements and widefield calcium imaging\cite{ren2021characterizing} for simultaneously capturing activity across all dorsal cortical areas -- allow large-scale monitoring at multiple spatial and temporal scales. Additional modalities, including behavior tracking\cite{mathis2018deeplabcut, syeda2024facemap}, cell-type labeling\cite{bugeon2022transcriptomic}, and connectivity mapping\cite{microns2025functional}, can further enhance forecasting performance and expand the range of scientific questions that forecasting can address. Finally, the abundance of publicly available mouse neural datasets\cite{de2023sharing,harris2019data} provides a unique opportunity to develop large-scale foundation models\cite{liang2024foundation} for neural activity forecasting in mice.

% The first and the only, at the time of writing and to the best of our knowledge, benchmark on neural time series forecasting was presented in \cite{lueckmann2025zapbench}, which compared linear and four nonlinear models for forecasting calcium activity recorded from a larval zebrafish brain. It was observed that when long history context is employed to generate forecasts, nonlinear deep learning based models generally outperform the linear model. Similarly, \cite{pankka2025enhanced} reported improved forecasts of resting-state EEG signals using deep neural networks compared to traditional linear autoregressive models. 
%!TEX root = ../main.tex

\section{Related Work}
Forecasting has long been an important approach in studying neural dynamics. 
Many earlier models, while not explicitly trained to optimize forecasting accuracy, are generative in nature and therefore can produce one-step-ahead or multi-step-ahead forecasts in an autoregressive fashion (e.g. \cite{curto2009simple,glaser2020recurrent,sani2021modeling}). As forecasting was only a secondary objective in these works, evaluations were generally restricted to a fixed prediction horizon, without systematic assessment across multiple horizons, and lack comparisons against forecasting baselines, such as the average of past activity or models capable of making direct-multi-step forecasts. Another major line of forecasting studies in neuroscience is motivated by brain-machine interface applications, where predicting neural activity is closely tied to predicting behavior. In this setting, models have been trained to solely predict behavior \cite{azabou2023unified,ryoo2025generalizable,azabou2025multisession} or to jointly predict neural activity and behavior \cite{sani2024dissociative,vahidi2025braid}. However, since such experiments often involve highly structured tasks (e.g., a monkey reaching to target), the resulting neural dynamics tend to be stereotyped (e.g., with rotational structure\cite{pandarinath2018inferring}), making them easier to forecast compared to less structured scenarios such as spontaneous activity. Forecasting accuracy has also been combined with other performance measures for both training and evaluation, for example in \cite{pei2021neural,zhang2024towards}.

More recently, a growing number of models have begun explicitly adopting forecasting accuracy as their primary training objective. For example, \cite{li2023amag} proposed a graph neural network based model for multi-channel neural activity forecasting; \cite{wagenmaker2024active} used a low-rank linear autoregressive model to predict neural responses to holographic photostimulation; \cite{antoniades2024neuroformer} developed a transformer-based model leveraging multi-modal inputs to autoregressively predict neural responses to visual stimuli; \cite{filipe2025one} introduced a diffusion-based model for joint forecasting of neural activity and behavior across sessions and subjects; \cite{immer2025forecasting} proposed a convolutional neural network based video model to directly forecast future frames of neural imaging videos; and \cite{duan2025poco} demonstrated forecasting of spontaneous neural activity across multiple sessions and subjects with a forecaster in the form of multilayer perceptron. While presented as neuroscience-specific applications, these models are fundamentally instances of general time series forecasting methods. Nevertheless, only a few (\cite{immer2025forecasting, duan2025poco}) have compared their performance against models from the broader time series forecasting literature, and even in those cases, the comparisons were limited. For example no recent foundation models (e.g., \cite{ansari2024chronos,woo2024unified}) were included. Moreover, all of these models produce only point forecasts. However, given the measurement and intrinsic noise in neural data \cite{faisal2008noise}, probabilistic forecasting \cite{gneiting2014probabilistic} -- which provides prediction intervals to quantify uncertainty -- is particularly important, but remains unexplored in the context of neural time series forecasting.

To fill these gaps, we systematically benchmarked common baseline methods, classical statistical approaches, and state-of-the-art deep learning models from the probabilistic time series forecasting literature on spontaneous mice neural activity. By drawing on the broader forecasting literature, we aim to identify strong model backbones that future neuroscience-specific applications can build upon for more accurate and uncertainty-aware neural activity predictions.

\begin{figure}[t!]
% \vspace{-24pt}
  \centering
\includegraphics[width=1.0\textwidth]{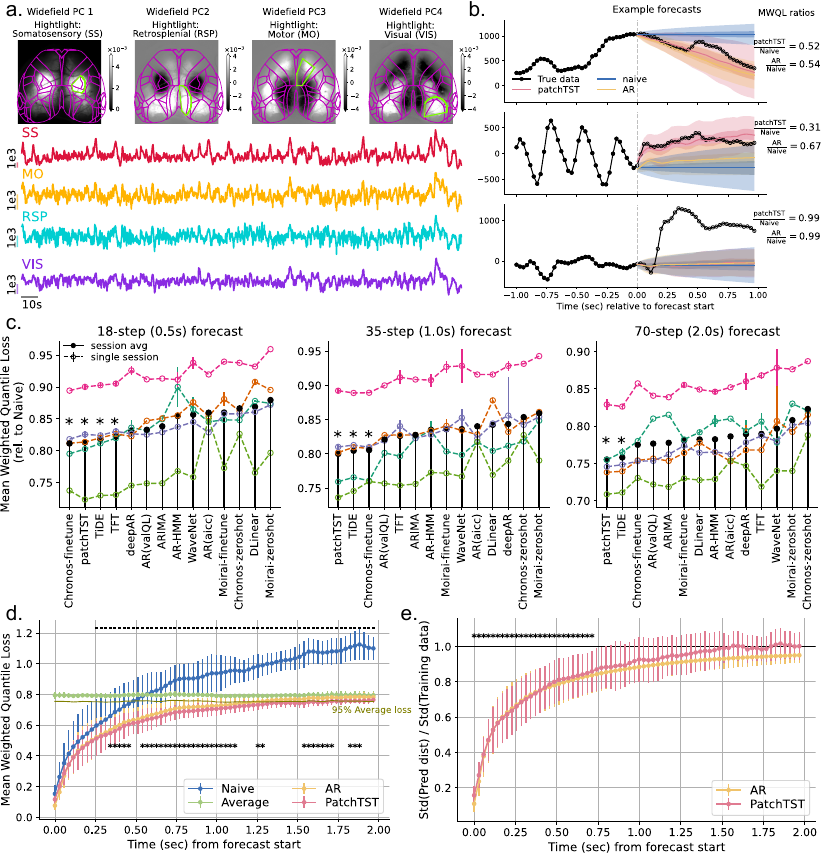}
  % \vskip -0.15cm
  \caption{\small \textbf{a.} Widefield imaging data aligned to Allen CCF, with example activity traces from one session.  \textbf{b.} True and predicted activity in three test samples; dashed lines mark forecast start. Shaded regions: prediction intervals (P.I.): dark = 20\%, light = 60\%. Top: PatchTST and AR perform similarly, both outperforming Naive. Middle: PatchTST outperforms both AR and Naive. Bottom: PatchTST and AR perform comparably to Naive. Quantitative performance in each sample is listed on the right. \textbf{c.} Model performance at three forecast horizons. Models sorted by mean performance across sessions. The y-axis indicates relative performance, computed as the Mean Weighted Quantile Loss (MWQL, see Appendix\ref{Appendix: models} for definition) across all predicted steps of each model divided by that of the Naive baseline (i.e., y=1 corresponds to Naive performance). Colors indicate different sessions. Errorbar: mean $\pm$ std across 5 random seeds; fine-tuning Chronos yielded identical results across random seeds, likely due to an issue in the AutoGluon implementation. Performances of Theta and Average are omitted, as they are close to Naive and worse than all other models shown; for 70-step forecast, Average outperforms Naive but still not the others. AR(aicc) vs. AR(valQL): order chosen by AICC vs. validation MWQL.
  Models with stars significantly outperform AR(valQL) (one-sided paired t-test, $p < 0.05$). \textbf{d.} MWQL by forecast step. Errorbar: mean $\pm$ std across sessions. Bars and stars indicate steps where PatchTST significantly outperforms Naive and AR, respectively (one-sided paired t-test, $p < 0.05$). PatchTST and AR losses exceed 95\% of the Average model loss (solid line) after 1.80s and 1.28s, respectively. \textbf{e.} Ratio of predicted distribution standard deviation (std) to training data std across forecast steps. Errorbar: mean $\pm$ std across sessions. Stars indicate steps where the PatchTST ratio is significantly less than 1 (one-sided t-test, $p<0.05$). In \textbf{b}, \textbf{d}, \textbf{e}, AR(valQL) is used as AR. For PatchTST, in each session the model achieving median MWQL across 5 random seeds is used.} 
  % \caption*{\footnotesize Note: Fine-tuning Chronos yielded identical results across random seeds, likely due to an issue in the AutoGluon implementation.}
\label{fig:fig1}
  % \vspace{-40pt}
  % \vskip -0.1cm
\end{figure}

\section{Results}

We systematically evaluated 12 univariate probabilistic forecasting models and 2 baseline methods on spontaneous neural activity in mouse cortex recorded using widefield calcium imaging\cite{ren2021characterizing} from five experimental sessions (details in Appendix\ref{Appendix: data}). Such data can be combined with cortex-wide optogenetic perturbations\cite{matveev2024simultaneous} to test forecast-driven closed-loop control. Widefield imaging was performed at 35 Hz, yielding recordings of spontaneous activity lasting $24.5 \pm 3.7$ minutes (corresponding to $51{,}495 \pm 7{,}784$ timesteps, mean $\pm$ std across five sessions). Data were registered to the Allen Mouse Brain Common Coordinate Framework (CCFv3)\cite{wang2020allen}, and we extracted average activity traces from four major brain regions: somatosensory (SS), motor (MO), visual (VIS), and retrosplenial (RSP) (Figure \ref{fig:fig1}a). 

For baselines, we consider the Naive method which repeats the last observed activity, and the Average method which takes the mean of past observed activity as prediction. Among the 12 models, 4 are classical statistical methods: the autoregressive (AR) model, autoregressive integrated moving average (ARIMA)\cite{shumway2006time}, autoregressive hidden Markov model (AR-HMM)\cite{kim2017state}, and the Theta model\cite{assimakopoulos2000theta}; 6 are deep learning based models: DeepAR\cite{salinas2020deepar}, DLinear\cite{zeng2023transformers}, Temporal Fusion Transformer (TFT)\cite{lim2021temporal}, PatchTST\cite{nie2022time}, Time-series Dense Encoder (TiDE)\cite{das2023long}, and WaveNet\cite{van2016wavenet}; and 2 are time series foundation models\cite{liang2024foundation}: Chronos\cite{ansari2024chronos} and Moirai\cite{woo2024unified}, for which we evaluated both zero-shot and fine-tuned performance. More details on models and training are provided in Appendix\ref{Appendix: models}, \ref{Appendix: training}.

The forecasting task was defined as predicting activity in the interval $[i, i+L)$ from the preceding observations in $[i-H, i)$, where $L$ is the forecast horizon and $H$ is the history length. Following previous forecasting literature\cite{salinas2020deepar,lim2021temporal, hyndman2021fpp3}, we partitioned all time series chronologically, using the earliest 60\% of timesteps for training ($\halfopen{0}{T_\text{train}}$), the next 20\% for validation ($\halfopen{T_\text{train}}{T_\text{val}}$), and the final 20\% for testing ($\halfopen{T_\text{val}}{T_\text{test}}$). Validation and test samples were generated using sliding windows with non-overlapping targets. For instance, the first test sample forecasts $\halfopen{T_\text{test}}{T_\text{test}+L}$ from $\halfopen{T_\text{test}-H}{T_\text{test}}$, the second forecasts $\halfopen{T_\text{test}+L}{T_\text{test}+2L}$ from $\halfopen{T_\text{test}+L-H}{T_\text{test}+L}$, and so on. In this way the forecast targets for evaluation remain fixed while varying the history length. We experimented with $L=18, 35, \text{and } 70$, which correspond to about $0.5, 1, \text{and } 2$ seconds. $H$ is treated as a hyperparameter and is tuned for each $L$ and each model individually.

In aggregated performance over test samples, all methods outperformed the Naive and Average controls (Figure \ref{fig:fig1}c, also see Appendix\ref{Appendix: suppfigures}, Figure \ref{fig:suppfig1} for evaluation on additional metrics). Classical autoregressive models provided competitive baselines, but several deep learning models -- PatchTST, TiDE, and fine-tuned Chronos -- consistently achieved higher accuracy across different prediction horizons. The strong performance of PatchTST aligns with findings from a recent benchmarking study on time series from various domains, such as economics, energy, and retail\cite{aksu2024gift}. In contrast, foundation models (Chronos and Moirai) pretrained on these domains transferred poorly to neural activity in a zero-shot setting, indicating that neural time series exhibit distinct characteristics from the data used for pretraining. Nevertheless, Chronos becomes competitive after fine-tuning on neural activity, suggesting that its architecture is sufficiently flexible to capture neural dynamics. At the level of individual test samples, we found that while some features of activity were captured by the models, there also exist large fluctuations that were not predicted by any model (Figure \ref{fig:fig1}b). 

The improved performance of models over controls extended across a range of forecast steps, out to more than 1 second. Note that in Figure \ref{fig:fig1}c, model performance is aggregated across all predicted steps. Therefore, it is actually unclear whether models still outperform baselines specifically at step 70, as the advantage may arise from better performance at earlier steps. When ploting error as a function of prediction step (Figure \ref{fig:fig1}d), we observed that performance of Naive, AR, and PatchTST all deteriorates with increasing horizon. While the Naive model continues to worsen, AR and PatchTST converge toward the Average model’s score. This behavior is expected for AR models: it can be verified that for stationary time series, $h$-step-ahead AR forecast will converge to the mean as $h \to \infty$. A similar pattern can be found in other metrics (Appendix\ref{Appendix: suppfigures}, Figure \ref{fig:suppfig2}). As PatchTST predicts all future steps simultaneously rather than autoregressively, separate models are optimized for each forecast horizon (18, 35, and 70 steps). We verified that the step-wise error curves are consistent across these models: for example, the first 18 steps of the 35- and 70-step models closely overlap with the 18-step model (Appendix\ref{Appendix: suppfigures}, Figure \ref{fig:suppfig3}). 

In addition, we quantified the uncertainty of probabilistic forecasts across steps, computed as the standard deviation of the predicted distribution over that of the training data (Figure \ref{fig:fig1}e). For both AR and PatchTST, this ratio of standard deviations increases with forecast horizon and approaches one (which is again expected for the AR model). Importantly, while wider prediction intervals (as suggested by larger ratio here) may be less informative for encompassing too many possibilities, intervals that are narrower are not necessarily better, as they may miss sources of uncertainty\cite{chatfield2000time}. Figure \ref{fig:fig1}d, e suggest that forecasts are most reliable within 35 steps (1 sec), and informative predictions may extend to about 50 steps (1.5 sec); beyond this, even the best of the current models (PatchTST) performs no better than simply predicting the mean and standard deviation of training data. 

\section{Discussion}
Several recent studies have proposed models tailored to forecasting neural time series. Nevertheless, models developed in the broader forecasting literature have not been systematically evaluated on neural activity, and the important aspect of uncertainty quantification has been largely overlooked. Here we bridged this gap by benchmarking models ranging from classical autoregressive methods to modern deep learning and foundation models on widefield imaging data. Several deep learning models consistently outperformed the strong baseline set by classical statistical models, yet both accuracy and uncertainty estimates indicate that current forecasts may only be informative up to a certain horizon. Whether this limit comes from model design constraints of existing methods or instead reflects intrinsic sources of variability and time scales in neural activity remains an open question. To probe the former, it may be of interest to develop neural time series forecasting foundation models by combining strong backbones identified here with neuroscience-specific innovations, such as cross-subject training\cite{azabou2023unified, azabou2025multisession, duan2025poco, filipe2025one}. Meanwhile, closed-loop control experiments informed by forecasting performance may help reveal the temporal structure inherent to neural system. 

\section*{Acknowledgments}
We gratefully acknowledge support from the National Science Foundation (NSF) (Research Grant 2024364 to ESB and NAS; CAREER award 2142911 to NAS). The work of ZL and JNK was supported in part by the NSF AI Institute for Dynamical Systems (dynamicsai.org), grant 2112085. Additional support was provided by the Pew Biomedical Scholars Program (to NAS), the Klingenstein-Simons Fellowship in Neuroscience (to NAS), the NIH Ruth Kirchenstein award (T32EY007031 and F31EY035880 to AJL), and
the Simons Foundation Shenoy Undergraduate Research Fellowship (to PM). We thank Kimberly Miller for help with mouse breeding and husbandry.

\bibliographystyle{unsrt}
\bibliography{ref.bib}

\newpage
\appendix
%!TEX root = ../main.tex

\section{Appendix}
\subsection{Data collection and preprocessing}\label{Appendix: data}
All experimental protocols were conducted according to US National Institutes of Health guidelines for animal research and approved by the Institutional Animal Care and Use Committee at the University of Washington.

We analyzed five sessions of widefield imaging experiments collected from five male and female mice from multiple GCaMP-expressing genotypes: two expressed GCaMP8s (tetO-GCaMP8s), one expressed GCaMP6s (tetO-GCaMP6s), and two expressed jGCaMP7f (Ai210 and Ai210 triple). Some of the data has been used in \cite{matveev2024simultaneous} or \cite{Ye2023.12.07.570517}, and the readers should refer to them for more experimental details. Briefly, images were collected with a CMOS camera with 17.3 um/pixel resolution (Basler acA2440-75m) and $560\times 560$ pixel frame size. The camera was fitted with a 0.63x objective lens (Leica Planapo 0.63x). The true frame rate is 70 Hz, but the effective rate is 35 Hz due to alternating 405 nm and 470 nm excitation for hemodynamic correction. Mice were head-fixed facing a dark screen, and were free to turn a wheel. In two of the five sessions, 10\% sucrose rewards were given in random 2-5 seconds intervals
to keep
mice at an awake and alert state. Three of the five sessions were solely spontaneous activity. In the other two sessions, receptive field mapping experiments were performed prior to spontaneous activity recording, and we analyzed spontaneous activity starting at least 2 minutes after the receptive field mapping experiments were turned off. Non-neuronal signals arising from hemodynamic changes were removed by subtracting the violet-light
evoked signals from the blue-light evoked signals with linear regression, as previously described in
\cite{zatka2021sensory, Ye2023.12.07.570517}. To store and process the data, we compressed the widefield data
($D$) into spatial components ($U$) and temporal components ($SV^\top$) with singular value decomposition in
the form $D = USV^\top$. For each of the four brain regions (SS, MO, RSP, VIS), activity was reconstructed using the top 500 (four sessions) or 200 (one session) SVD components and was averaged across pixels within that region. It was reported in \cite{matveev2024simultaneous} and \cite{Ye2023.12.07.570517} that over 97\% of the total variance before SVD compression was captured within the top 50 components. As we have used hundreds of components here, the information lost in compression and reconstruction should be negligible. 

% ZYE69: F, tetO-GCaMP6s, No reward
% AL33: M, tetO-GCaMP8s, with reward
% AL39: ?, tetO-GCaMP8s, with reward
% ZYE43: F, Ai210,php.eb-cre, No reward
% LK03: M, Ai210 triple, No reward

\subsection{Univariate time series models}\label{Appendix: models}
\subsubsection{Baselines \cite{hyndman2022variance}}
\begin{enumerate}
    \item \textbf{Naive (repeat-last-step).} Let $y_t$ denote the time series at timestep $t$, and $\{\epsilon_t\}$ be white noise with mean 0 and variance $\sigma^2$. The Naive baseline assumes $y_t$ follows a random walk: $y_t = y_{t-1} + \epsilon_t$. Thus, to forecast from $y_T$, we have 
    \begin{align*}
        y_{T+1} &= y_T + \epsilon_{T+1} \\
        y_{T+2} &= y_{T+1} + \epsilon_{T+2} = y_T + \epsilon_{T+1} + \epsilon_{T+2} \\
        \hdots \\
        y_{T+h} &= y_T + \sum_{i=1}^h \epsilon_{T+i}
    \end{align*}
    Let $y_{T+h|T}$ denote the forecast of $y_{T+h}$ using previous observations up to (and including) $y_{T}$. It follows that 
    \begin{align*}
        \hat{y}_{T+h|T} & \vcentcolon= \mathbb{E}[y_{T+h|T}] = y_T, \\
        \hat{\sigma}_h^2 & \vcentcolon= \Var(y_{T+h|T}) = \Var({\sum_{i=1}^h \epsilon_{T+i}}) = \sum_{i=1}^h \Var(\epsilon_{T+i}) = h\sigma^2,
    \end{align*}
    where $\sigma^2$ can be estimated from fitted residuals: 
    \[
    \sigma^2 = \frac{1}{T-1}\sum_{i=2}^T (y_i - y_{i|i-1})^2 = \frac{1}{T-1}\sum_{i=2}^T (y_i - y_{i-1})^2.
    \]
    Assuming $\epsilon_{t}$ is normal, then a $100(1-\alpha)\%$ prediction interval is 
    \[
    \left[ \hat{y}_{T+h|T} - z_{\frac{\alpha}{2}}\hat{\sigma}_h, \hat{y}_{T+h|T} + z_{\frac{\alpha}{2}}\hat{\sigma}_h \right],
    \]
    where $z_{\frac{\alpha}{2}}$ is the value above which a fraction of $\frac{\alpha}{2}$ of the data in a standard normal distribution falls.

    \item \textbf{Average.} Again let $y_t$ denote the time series at timestep $t$, and $\{\epsilon_t\}$ be white noise with mean 0 and variance $\sigma^2$. The Average baseline assumes $y_t$ randomly fluctuates around some mean value: $y_t = c + \epsilon_t$, where $c$ is some constant to be estimated from the past observations. To forecast from $y_T$, we first compute the least-squares estimate of $c$: 
    \[
    \hat{c} = \frac{1}{T}\sum_{j=0}^{T-1} y_{T-j}, 
    \] then for $i=1, 2, \dots$, 
    \[
    y_{T+i} = \hat{c} + \epsilon_{T+i}. 
    \]
    It follows that 
    \begin{align*}
        \hat{y}_{T+h|T} & \vcentcolon= \mathbb{E}[y_{T+h|T}] = \hat{c}, \\
        \hat{\sigma}_h^2 & \vcentcolon= \Var(y_{T+h|T}) = \Var(\hat{c}) + \Var(\epsilon_{T+i}) = \frac{1}{T}\sigma^2 + \sigma^2,
    \end{align*}
    where $\sigma^2$ can be estimated from fitted residuals: 
    \[
    \sigma^2 = \frac{1}{T-1}\sum_{j=0}^{T-1} (y_{T-j} - \hat{c})^2.
    \]
    The prediction interval can be constructed in the same manner as in the Naive method. Note that $\hat{\sigma}_h$ is actually independent of $h$, and thus the width of prediction interval is constant through all forecasting steps.
    
    \end{enumerate}

\subsubsection{Local models}
The following models are ``local" in the sense that we need to fit separate models to each series if there are multiple time series in one dataset. 
    \begin{enumerate}
    \item \textbf{Autoregressive integrated moving average (ARIMA) family \cite{shumway2006time}.} Let $y_t$ denote the time series at timestep $t$, and $\{\epsilon_t\}$ be white noise. An ARIMA($p, q, d$) model is of the form 
    \[
    y'_t = c + \phi_1 y'_{t-1} + \cdots + \phi_p y'_{t-p} + \theta_1 \epsilon_{t-1} + \cdots + \theta_q \epsilon_{t-q} + \epsilon_{t}
    \]
    where $y'_t$ is a $d$-th order differenced version of $y_t$. We applied both the KPSS test and the augmented Dickey–Fuller test to each time series here and found that they all satisfy the stationarity condition. So we take $d=0$, $y'_t = y_t$, and the ARIMA model simplifies to an ARMA model. The \textbf{autoregressive (AR) model} can be seen as a special case of ARIMA, where $d=0$, and the $\theta's$ are taken to be zero. During fitting, for fixed $p, q$, the parameters $c, \phi_i, \theta_i$ are determined by maximizing the likelihood of observed data under the assumption that $\{\epsilon_t\}$ is Gaussian
    \[
    L(c, \{\phi_i\}_{i=1}^p, \{\theta_i\}_{i=1}^q) = \prod_{t=1}^n f(y'_t|y'_{t-1}, \dots, y'_1, c, \{\phi_i\}_{i=1}^p, \{\theta_i\}_{i=1}^q).
    \]
    To find the most appropriate value of $(p, q)$, a common approach is to compute the likelihood for various values of $(p, q)$, and choose the pair achieving the minimum AICC\cite{brockwell2002introduction}. Alternatively, we can evaluate fitted models on a validation set, and selected the $(p, q)$ pair achieving the best score on the validation set. \\
    The \textbf{autoregressive hidden Markov model (AR-HMM)} extends the AR model, allowing the parameters to switch among different values. It can be written as
    \[
    y_t = c^{S_t} + \phi^{S_t}_1 y_{t-1} + \cdots + \phi^{S_t}_p y_{t-p} + \epsilon^{S_t}_{t}, 
    \]
    where $S_t$ denotes the state (assumed to be hidden, unobserved) at timestep $t$, and $\epsilon^{S_t}_{t} \sim \mathcal{N}(0, (\sigma^{S_t})^2)$. $S_t$ switches among a set of discrete values following a first-order Markov chain. Since it is possible to transition into several different states at each step (although some states are more probable than others), AR-HMM can make probabilistic forecasts by sampling multiple trajectories from the forecast start. 

    \item \textbf{Theta family \cite{assimakopoulos2000theta, fiorucci2016models}.} The general idea of Theta model is to decompose the time series into multiple components with different curvatures (as controlled by second derivatives), extrapolate each component separately into the future, and then combine their extrapolations. In its most basic form proposed in \cite{assimakopoulos2000theta}, two Theta lines are drawn from the original time series: one with second derivative equal to 0 ($\Theta = 0$) for extracting long-term linear trend, and the other with second derivative double that of the original series ($\Theta = 2$) for extracting local fluctuations. During prediction, the first Theta line ($\Theta = 0$) continues to follow the linear trend, and the second Theta line ($\Theta = 2$) is extrapolated via simple exponential smoothing. A simple average of their extrapolations is taken to be the final forecast. Methods to further improve $\Theta$-parameter selection and extrapolations were later proposed in \cite{fiorucci2016models}. In a recent benchmark study \cite{aksu2024gift}, the Theta family demonstrates competitive performance on time series with high sampling frequency (10 sec). Nonetheless, as those series are from a very different domain (cloud operations) than neuroscience, it is unclear whether the Theta family is also effective for forecasting widefield imaging data. 
    
    \end{enumerate}

\subsubsection{Global models}
The following models are ``global" in the sense that one model can simultaneous fit all time series in a dataset. 
    \begin{enumerate}
    \item \textbf{DeepAR\cite{salinas2020deepar}.} DeepAR model makes probabilistic forecasts with an autoregressive recurrent neural network (RNN). The value of the time series at the next timestep is assumed to be drawn from some probability distribution, which is usually taken to be Gaussian or Student-t for continuous data. The parameters of the distribution (e.g., the mean and variance for Gaussian distribution) are determined by the output of a multi-layer RNN: Let $y_t$ denote time series observations and $x_t$ denote some known covariate. At timestep $t$, the RNN computes $\vh_t$ from $\vh_{t-1}, y_{t-1}, x_t$, and assumes $y_t \sim \theta(\vh_t)$, where $\theta(\cdot)$ is some fixed distribution. In practice, it is often advantageous to also incorporate lagged versions of $y_{t-1}$ and $x_t$ (e.g. $y_{t-2}, x_{t-1}$) when computing $\vh_t$. 
    % The model is trained to maximize the joint log-likelihood across all time series and all time steps. The model is trained to maximize the joint log-likelihood across all time series (as indexed by $i$) and all time steps (as indexed by $t$): $L = \sum_{i=1}^N \sum_{t = 0}^T \log l(y_{i, t}|\theta(\vh_{i, t}))$. When the relevant history context length $H$ and the forecasting horizon $L$ are specified, if the total observations of the time series $T$ is greater than $H+L$, then multiple training instances will be constructed using sliding windows. 
    During inference, to forecast $L$ steps from $t=T+1$, the model will first run from $t=T-H$ (where $H$ is the relevant history context length) to $T$ to generate $\vh_{T+1}$, sample one $\hat{y}_{T+1}$ from $\theta(\vh_{T+1})$, and then use $\hat{y}_{T+1}, \vh_{T+1}, x_{T+1}$ to compute $\vh_{T+2}$ and generate a sample trajectory in this autoregressive form. To make probabilistic forecasts, multiple sample trajectories will be generated, and the prediction interval at each time step will be estimated from the sample trajectories. 
    
    \item \textbf{DLinear\cite{zeng2023transformers}.} DLinear model is a one-layer linear feedforward network for direct multi-step forecasting. It was originally proposed to showcase that the advantage of many transformer-based models in time series forecasting mainly comes from the direct multi-step-ahead forecasting strategy (as opposed to autoregressive one-step-ahead), instead of the transformer architecture. Suppose we want to forecast $\vy^\text{future} = \left[y_T, \dots, y_{T+L}\right]$ from $\vy^\text{history} = \left[y_{T-H}, \dots, y_{T}\right]$. Using a moving average smoother, $\vy^\text{history}$ is first decomposed into a trend and a remainder component: $\vy^\text{history} = \vy^\text{history, trend} + \vy^\text{history, remainder}$. Each of the two components is passed through a one-layer linear feedforward network, and then summed together as the future forecast: $\hat{\vy}^\text{future} = \mW_1 \vy^\text{history, trend} + \mW_2 \vy^\text{history, remainder}$, where $\mW_1, \mW_2$ are $L\times H$ matrices. If the time series is multivariate, i.e., for the $i$-th variate, $\vy^{(i), \text{future}} = \left[y_T^{(i)}, \dots, y_{T+L}^{(i)}\right]$, $\vy^{(i), \text{history}} = \left[y_{T-H}^{(i)}, \dots, y_{T}^{(i)}\right]$, then $\mW_1, \mW_2$ will be shared across all variates and no cross-variate relationship will be modeled, i.e., $\hat{\vy}^{(i), \text{future}} = \mW_1 \vy^{(i), \text{history, trend}} + \mW_2 \vy^{(i), \text{history, remainder}}$ for all $i$. 
    
    The original DLinear model only produces point forecasts, but the GluonTS package\cite{gluonts_jmlr} adds an additional simple transformation step to make it output the parameters of the desired distribution and thus make the forecasts probabilistic. For example, for the Gaussian distribution, $\hat{y}_t^\text{future} \sim \mathcal{N}(\mu(\vy^\text{feature}), \sigma(\vy^\text{feature}) )$, where $\vy^\text{feature} \in \sR^D$. $\vy^\text{feature} = \mW_1 \vy^\text{history, trend} + \mW_2 \vy^\text{history, remainder}$ as in the original DLinear model. $\mu(\vy^\text{feature}) = \vw_{\mu}^\top \vy^\text{feature} + \vb_{\mu}$, and $\sigma(\vy^\text{feature}) = \log(1+\exp(\vw_{\sigma}^\top \vy^\text{feature} + \vb_{\sigma}))$.
    
    \item \textbf{TFT\cite{lim2021temporal}.} Temporal fusion transformer (TFT) is a versatile model that can incorporate static covariates, covariates that are only known in the past, and covariates that are known for the forecasting horizon. It first uses recurrent layers to extract local temporal information that is shared across neighboring time steps, and then uses the attention mechanism to process global temporal information that are spread across a wider time frame. 
    It performs probabilistic forecasting by optimizing the quantile loss. However, one caveat is that it does not address the problem of quantile crossing\cite{park2022learning}. For instance, the predicted 0.5-quantile may be larger than the predicted 0.9-quantile.  
    
    \item \textbf{PatchTST\cite{nie2022time}.} PatchTST model uses the vanilla transformer encoder\cite{vaswani2017attention} as backbone, together with two key designs: channel-independence and patching. Suppose we want to forecast a $M$-dimensional multivariate time series $\vy^\text{future} = \left[\vy_T, \dots, \vy_{T+L}\right]$ from $\vy^\text{history} = \left[\vy_{T-H}, \dots, \vy_{T}\right]$, where $\vy_t \in \sR^{M}$. PatchTST models each variate individually using a shared model $\mathcal{F}$ (``channel-independence"): $\vy^{(i), \text{future}}= \mathcal{F}(\vy^{(i), \text{history}})$, where $\vy^{(i), \text{future}} = \left[y^{(i)}_T, \dots, y^{(i)}_{T+L}\right], \vy^{(i), \text{history}} = \left[y^{(i)}_{T-H}, \dots, y^{(i)}_{T}\right]$. For each variate $i$, neighboring timesteps are grouped together as one ``patch" (or token), and then fed into the transformer encoder. The patches may or may not be overlapping, and the number of patches is typically smaller than the number of history context timesteps. Each patch will be linearly mapped to a high-dimensional vector, and go through nonlinear transformations (such as self-attention and feed-forward networks) in the transformer encoder. $\hat{\vy}^{(i), \text{future}}$ is computed from linear transformation of the transformer encoder output. Similar to DLinear, the original PatchTST model only produces point forecasts, but is adapted for probabilistic forecasting by the GluonTS\cite{gluonts_jmlr} package.

    \item 
    \textbf{TiDE\cite{das2023long}.} Time-series Dense Encoder (TiDE) can be thought as a nonlinear extension of the DLinear model. While DLinear only accounts for the linear relationship, TiDE uses the classical multi-layer perceptron and residual connections to capture the potentially nonlinear relationship between future and observed activity. When there are multiple time series in the dataset, it operates in a channel-independent manner, as PatchTST, and does not model the relationship between different time series. Similar to DLinear and PatchTST, the original TiDE model only produces point forecasts, but is adapted for probabilistic forecasting by the GluonTS\cite{gluonts_jmlr} package. 

    \item \textbf{WaveNet\cite{van2016wavenet}.} WaveNet is a convolutional neural network-based autoregressive model. For a sequence $\left[y_1, \dots, y_T\right]$, it learns $p(y_t|y_{t-1}, \dots, y_1)$ for all $t$. WaveNet is originally proposed for modeling audio signals, which are usually stored as 16-bit integers so only take a finite number of values. Thus WaveNet can learn $p(y_t|\cdots)$ as categorical distributions and can be trained using the cross-entropy loss. GluonTS\cite{gluonts_jmlr} adapts WaveNet to model general time series by discretizing the real-valued observations into a fixed number of bins so they can be modeled as categorical distributions too. Similar to DeepAR, WaveNet generates probabilistic forecasts by autoregressively sampling different trajectories. In a recent study\cite{pankka2025enhanced}, WaveNet was applied to forecast resting-state EEG signal and was observed to outperform the classical AR model. 
    
    \end{enumerate}

\subsubsection{Foundation models}
    \begin{enumerate}
    \item \textbf{Chronos\cite{ansari2024chronos}.} Chronos is a large language model (LLM)-based method for forecasting univariate time series, using the T5 model\cite{raffel2020exploring} as backbone. The idea is that, upon proper tokenization, time series can be forecast in the same way as text using LLM. Since LLMs work with a finite dictionary of tokens (i.e., words), Chronos discretizes a real-valued time series into a fixed number of bins, and is trained using the cross-entropy loss (similar to the adapted WaveNet).
    % Chronos first converts a real-valued time series ($y_i \in \sR$) into a set of discrete tokens ($y'_i \in {1, \dots, B}$) via quantization: defined $B$ bins along the real line, let $y'_i = i$ if $y_i$ falls within the $i$-th bin. Both the true and predicted values are modeled as the categorical distribution, and the model is trained by minimizing the cross-entropy loss: $$l(\vtheta) = -\sum_{h=0}^H \sum_{i=1}^B \mathbbm{1}\{y'_{C+h+1=i}\} \log p_{\vtheta} (y'_{C+h+1=i}|y'_1, \dots, y'_{C+h}),$$ where $\vtheta$ denotes the parametrization to be learned by the model, $y'_1, \dots, y'_C$ are observed history values, and $y'_{C+1}, \dots, y'_{C+H}$ are forecast targets. 
    During prediction, the original Chronos model proposed in \cite{ansari2024chronos} follows the autoregressive one-step-ahead forecasting strategy, similar to DeepAR. The later version, Chronos-Bolt\cite{ansari2024chronosbolt}, adopts the direct multi-step-ahead strategy, which generates estimates of quantiles $0.1, \dots, 0.9$ of all predicted steps simultaneously using a feedforward network applied to decoder output. However, similar to TFT, Chronos-Bolt also faces the problem of quantile crossing\cite{park2022learning}. 
    
    As with other LLMs, Chronos has been trained on a huge amount of data. Its training set consists of 28 real-world datasets with about 890K univariate time series spanning multiple domains (weather, finance, transportation, etc). It is trained with maximum context length of 512 and maximum forecast horizon of 64 time steps. Because of its rich training history, Chronos may be capable to perform zero-shot forecasting on time series that have never been seen by the model before. Nonetheless, since the majority of the time series in the training set of Chronos is of low frequency (sampled at hourly or lower rate), it is unclear whether the pretraining of Chronos can actually benefit it in forecasting widefield data. 
    
    \item \textbf{Moirai\cite{woo2024unified}.} Moirai is another LLM-based time series forecasting model. Its main differences from Chronos are that it can handle time series with covariates and does not require discretizing time series into categorical variables. To achieve the former, it flattens the multivariate time series (e.g., a matrix of shape $d \times T$) into one sequence (e.g. a vector of length $d \times T$), while keeping track of the variate identities and time steps. Also, similar to PatchTST, instead of defining each time step as a separate token input to the underlying transformer model, Moirai forms ``patches" by grouping nearby time steps of the same variate, and take each patch as a token. The patch size is set based on the frequency of the data. To model time series with diverse distributions of values without mapping them to categorical variables, Moirai is trained by maximizing the log-likelihood with respect to a mixture of probability distributions: 
    \[
    l(\vtheta) = \log p(y_{T+1}, \dots, y_{T+h}|f_{\vtheta}(y_{T-C}, \dots, y_{T})), \text{with } p(\cdot) = \sum_{i=1}^4 w_i p_i(\cdot),
    \]
    where the $w_i's$ are also learnable parameters and the $p_i's$ are taken to be the Student's t, negative binomial, log-normal, and normal distributions. Probabilistic forecasts can be generated by sampling from this mixed distribution. Just like Chronos, Moirai has also been trained on a huge amount of data with a total of 27 billion observations, and thus may be capable of zero-shot forecasting. During training, Moirai uses samples with randomized history and prediction length to further improve its applicability to diverse forecasting problems. 
    
\end{enumerate}

\subsection{Metrics}\label{Appendix: metrics}
Let $y_{i, t}$ denote the true activity at $t$-th forecast step in the $i$-th test sample, $y_{i, t}\sim D_{i, t}$, $t=1, \dots, H, i=1, \dots, N$. Let $f^q_{i, t}$ denote the predicted $q$-th quantile of $D_{i, t}$. 

\begin{figure}[h!]
% \vspace{-20pt}
  \centering
\includegraphics[width=1.0\textwidth]{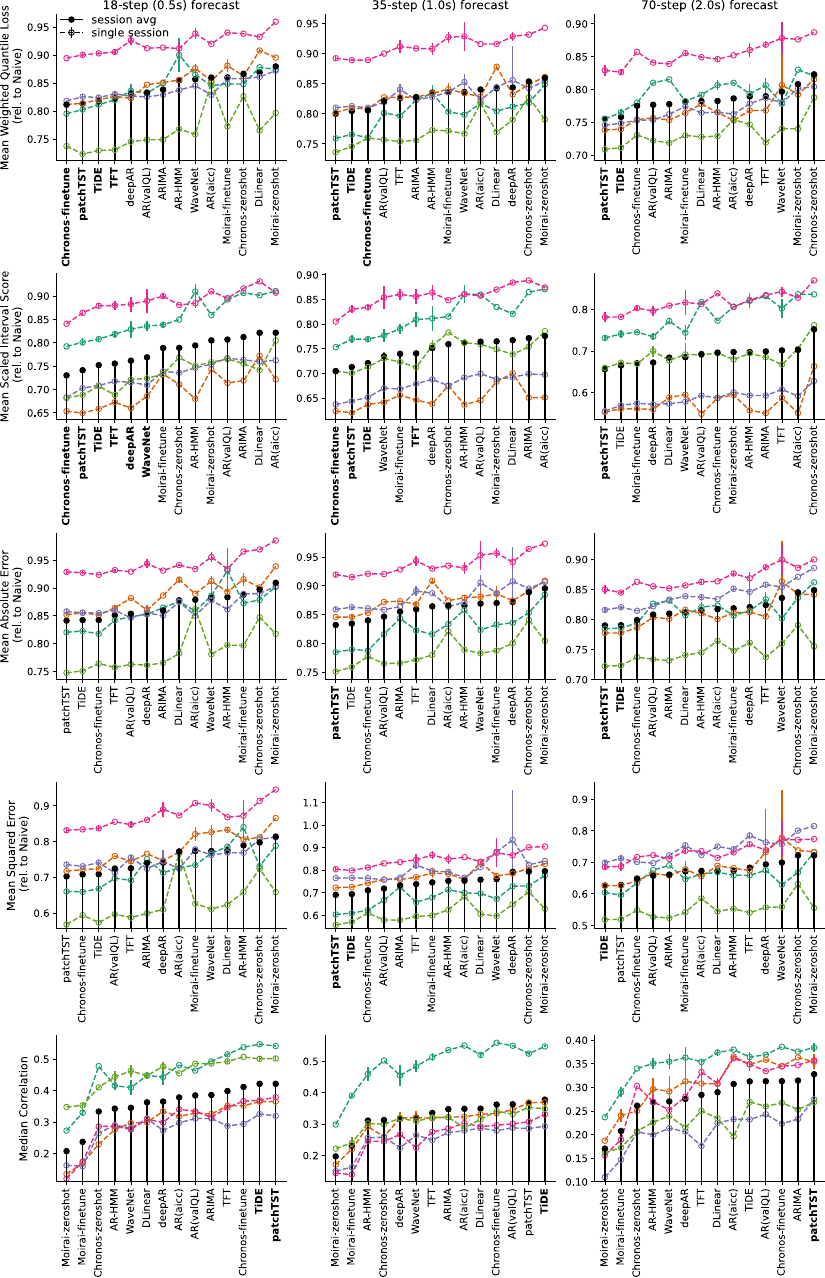}
  % \vskip -0.2cm
  \caption{\small Evaluation on additional metrics. Models sorted by mean performance across sessions. Models in boldface significantly outperform AR(valQL) (one-sided paired t-test, $p < 0.05$). For Chronos and Moirai, -FT indicates the finetuned version, -ZS indicates zeroshot version.} 
\label{fig:suppfig1}
  % \vspace{-10pt}
  % \vskip -0.2cm
\end{figure}

\begin{figure}[h!]
% \vspace{-20pt}
  \centering
\includegraphics[width=0.8\textwidth]{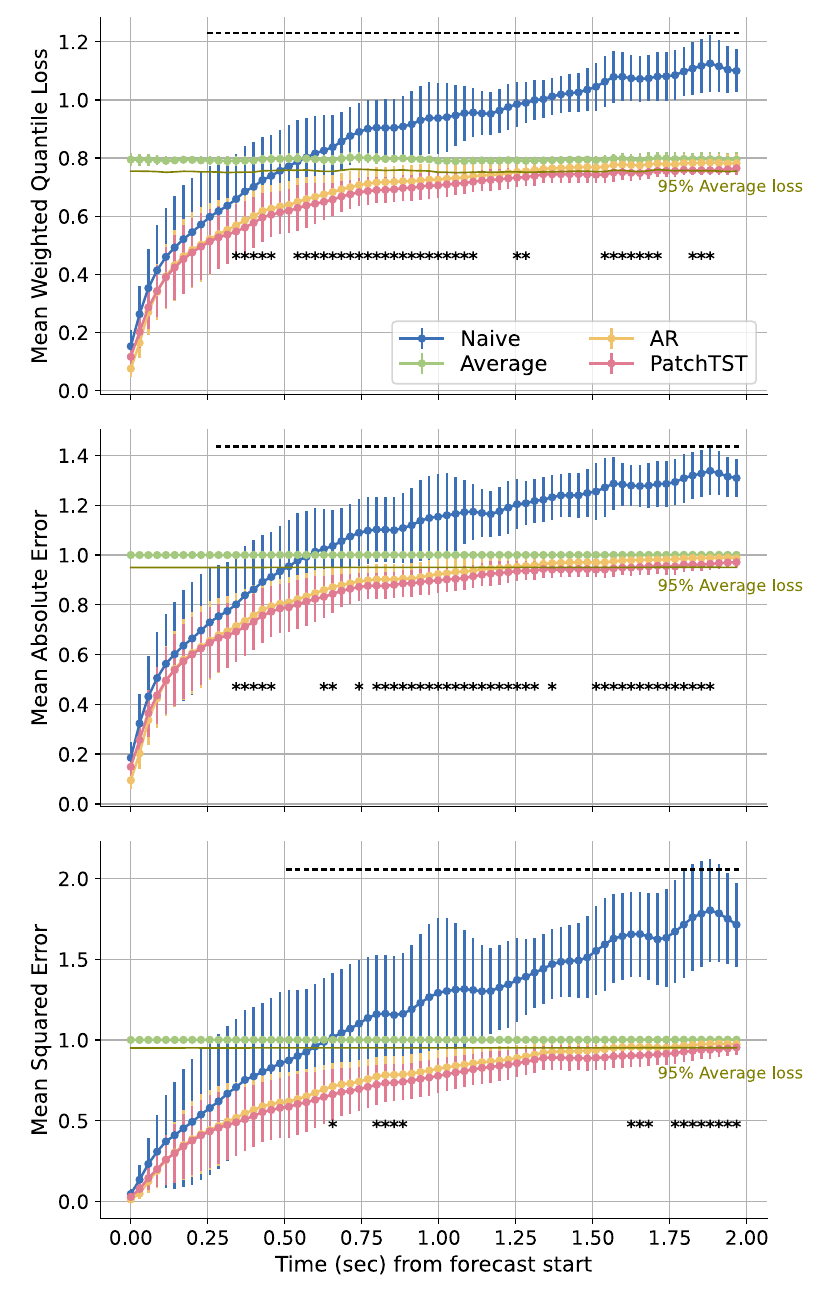}
  % \vskip -0.2cm
  \caption{\small Additional metrics over prediction steps. Bars and stars indicate steps where PatchTST significantly outperforms Naive and AR, respectively (one-sided t-test, $p < 0.05$). PatchTST loss exceeds 95\% of the Average model loss (solid line) after 1.80s (top), 1.65s (middle), 1.97s (bottom). AR loss exceeds 95\% of the Average model loss after 1.28s (top), 1.23s (middle), 1.60s (bottom). } 
\label{fig:suppfig2}
  % \vspace{-10pt}
  % \vskip -0.2cm
\end{figure}

\begin{figure}[h!]
% \vspace{-20pt}
  \centering
\includegraphics[width=0.8\textwidth]{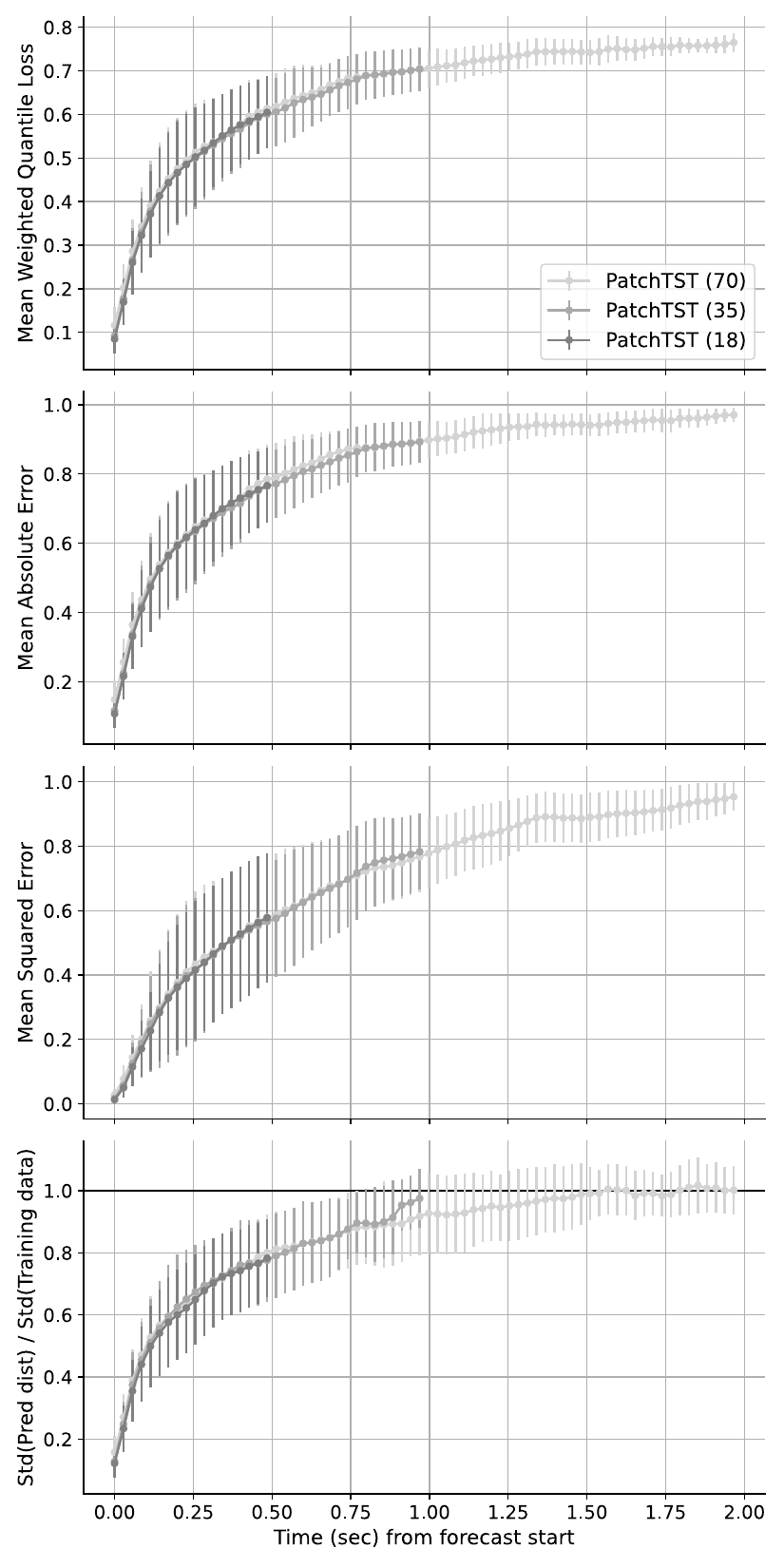}
  % \vskip -0.2cm
  \caption{\small Performance across PatchTST models trained with different forecast horizons. PatchTST (18), PatchTST (35), PatchTST (70): PatchTST trained with forecast horizon of 18, 35, and 70 steps.} 
\label{fig:suppfig3}
  % \vspace{-10pt}
  % \vskip -0.2cm
\end{figure}

\begin{enumerate}
\item \textbf{Mean Weighted Quantile Loss (MWQL):} 
When aggregating across all test samples and prediction steps (Figure \ref{fig:fig1}c), we computed
\[
\text{MWQL} = \frac{\sum_{i=1}^N \sum_{t=1}^{H} \frac{1}{Q} \sum_{q} \rho_q(y_{i, t}, f^q_{i, t})}{\sum_{i=1}^N \sum_{t=1}^{H} |y_{i, t}|}.
\]
When aggregating across all test samples for each prediction steps $t$ (Figure \ref{fig:fig1}d), we computed
\[
\text{MWQL}_t = \frac{\sum_{i=1}^N  \frac{1}{Q} \sum_{q} \rho_q(y_{i, t}, f^q_{i, t})}{\sum_{i=1}^N |y_{i, t}|}.
\]
In both cases, $q=0.1, 0.2, \dots, 0.9$, $Q=9$. $\rho_q$ is the quantile score\cite{hyndman2021fpp3}:
\begin{equation*}
    \rho_q(y_{i, t}, f^q_{i, t}) = \begin{cases} 2(1-q)(f^q_{i, t}-y_{i, t}), \; \text{if } y_{i, t} < f^q_{i, t} \\ 2q(y_{i, t}-f^q_{i, t}), \; \text{if } y_{i, t} \geq f^q_{i, t} \end{cases}
\end{equation*}
MWQL has been popularly used for evaluating probabilistic forecasts, e.g., in \cite{park2022learning, ansari2024chronos, woo2024unified}. 
\item \textbf{Mean Scaled Interval Score (MSIS):}
\[
\text{MSIS} = \frac{1}{N} \sum_{i=1}^N \frac{\sum_{t=1}^{H} W^\alpha_{i, t}(y_{i, t}, u^\alpha_{i, t}, l^\alpha_{i, t})}{\sum_{t=1}^{H} |y_{i, t}|}
\]
Here $[l^\alpha_{i, t}, u^\alpha_{i, t}]$ define a $100(1-\alpha)\%$ prediction interval for $y_{i, t}$, typically $l^\alpha_{i, t} = f^{\frac{\alpha}{2}}_{i, t}$, $u^\alpha_{i, t} = f^{1-\frac{\alpha}{2}}_{i, t}$.
$W^\alpha$ is the Winkler score\cite{hyndman2021fpp3}, which combines penalties for the width of the interval and for lack of coverage:
\[
W^\alpha (y_{i, t}, u^\alpha_{i, t}, l^\alpha_{i, t}) = (u^\alpha_{i, t}- l^\alpha_{i, t}) + \frac{2}{\alpha} (l^\alpha_{i, t} - y_{i, t})\mathbbm{1}(y_{i, t} < l^\alpha_{i, t}) + \frac{2}{\alpha} (y_{i, t}-u^\alpha_{i, t})\mathbbm{1}(y_{i, t} > u^\alpha_{i, t})
\]
In this paper we took $\alpha = 0.2$. \cite{park2022learning, woo2024unified} used a similar metric to evaluate prediction intervals, where the denominator is the absolute seasonal error. Since seasonality is not obvious in neural time series, we used the absolute target value instead. 

\item \textbf{Mean Absolute Error (MAE) and Mean Squared Error (MSE):}
When aggregating across all test samples and prediction steps (Figure \ref{fig:suppfig1}), we computed
\[
\text{MAE} = \frac{\sum_{i=1}^N \sum_{t=1}^{H} |y_{i, t} - \hat{y}_{i, t}|}{\sum_{i=1}^N \sum_{t=1}^{H} |y_{i, t}|}, \; \text{MSE} = \frac{\sum_{i=1}^N \sum_{t=1}^{H} |y_{i, t} - \hat{y}_{i, t}|^2}{\sum_{i=1}^N \sum_{t=1}^{H} |y_{i, t}|^2}
\]
When aggregating across all test samples for each prediction steps $t$ (Figures \ref{fig:suppfig2},\ref{fig:suppfig3}), we computed
\[
\text{MAE}_t = \frac{\sum_{i=1}^N |y_{i, t} - \hat{y}_{i, t}|}{\sum_{i=1}^N |y_{i, t}|}, \; \text{MSE}_t = \frac{\sum_{i=1}^N |y_{i, t} - \hat{y}_{i, t}|^2}{\sum_{i=1}^N |y_{i, t}|^2}
\]
Here $\hat{y}_{i, t}$ is a point forecast of $y_{i, t}$, and is taken to be the median value of probabilistic forecasts, i.e., $f^{0.5}_{i, t}$. MAE was also used in \cite{salinas2020deepar}, where it was called Normalized Deviation.

\item \textbf{Correlation:} We computed the Pearson's correlation coefficient $r$ between $y_i = \{y_{i, t}\}_{t=1,\dots, H}$ and $\hat{y}_{i} = \{\hat{y}_{i, t}\}_{t=1,\dots, H}$ for each test sample. The median $r$ across of all test samples is reported is Figure \ref{fig:suppfig1}.

\end{enumerate}

\subsection{Model implementation and training details}\label{Appendix: training}

% In each session, classical statistical models were fit separately to each region, while deep learning models were trained jointly across all regions. Model hyperparameters were chosen via either grid search or random search (details in Appendix\ref{Appendix: training}). For deep learning models, during training, samples were constructed by randomly selecting a timestep $i$ from the interval $\halfopen{H}{T_\text{train} - L}$. Each sample consisted of a context window $\halfopen{i-H}{i}$ and a forecast target $\halfopen{i}{i+L}$. 

For the \textbf{Naive} and \textbf{Average} baselines, we used the implementations from the StatsForecast package \cite{garza2022statsforecast}. These models have no trainable parameters or hyperparameters.

For the \textbf{AR} model, we considered two order–selection strategies:
\begin{enumerate}
    \item AICC-based (\textbf{AR(aicc)}): we used \texttt{AutoARIMA} from StatsForecast, which by default uses the Hyndman-Khandakar stepwise search algorithm to find the model order and parameter achieving the lowest AICC\cite{hyndman2021fpp3}. Models were fitted using both training and validation sets. We set the limit on the number of models to explore to 100. 

    \item Validation-based (\textbf{AR(valQL)}): we used \texttt{AutoRegressive} from StatsForecast. Starting from lag order 1, we fitted the model on the training set and computed its MWQL on the validation set. The lag order was increased by one until validation MWQL failed to improve for 10 consecutive orders; the best-performing model so far was then chosen and evaluated on the test set.
\end{enumerate}

For \textbf{ARIMA}, we again used \texttt{AutoARIMA} from StatsForecast. The AR order ($p$) was initialized with the optimal order found by \textbf{AR(aicc)}, and the MA order ($q$) was initialized to 0. Models were fitted using both training and validation sets. We set the limit on the number of models to explore to 100.

For \textbf{Theta}, we fitted four variants provided in StatsForecast (standard, optimized, dynamic standard, and dynamic optimized) on the training set. The variant with the lowest validation MWQL was evaluated on the test set.

For \textbf{AR-HMM}, we used \texttt{LinearAutoregressiveHMM} from the Dynamax package\cite{Linderman2025}. We fitted models with number of states $S \in \{2, 3, 4, 5, 6, 7, 8, 9, 10\}$ and number of lags $L \in \{1, 2, 4, 6, \dots, 40, 44, 48, \dots, 80
\}$ on the training set, and evaluated their log likelihoods of the validation data ($\text{LL}_\text{val}$). The optimal $S$ was chosen as the smallest value whose $\text{LL}_\text{val}$ was within 0.1\% of the maximum across all models (i.e., $\geq 1.001\times \text{max }\text{LL}_\text{val}$, since LLs are negative). The optimal $L$ was chosen as the smallest value whose $\text{LL}_\text{val}$ was within 0.1\% of the maximum across all lags at the selected $S$. We then fitted models with the optimal ($S, L$) with 5 random seeds on the training set and evaluated them on the test set. For each test sample, we generated 100 forecast trajectories to compute prediction intervals.

For \textbf{Global} models, we used the implementation from GluonTS\cite{gluonts_jmlr}. Some models (\textbf{DLinear}, \textbf{PatchTST}, \textbf{TiDE}) was originally proposed for point forecast, but were adapted to generate probabilistic forecasts by GluonTS. See the \textbf{DLinear} section in \ref{Appendix: models} for an example of such adaption. All models assumed Student-t output distributions. During training, all models employed early stopping based on validation loss, terminating if no improvement occurred for 10 epochs. Since widefield data does not clear seasonality, we disabled automatically added time features (e.g., hour of day, day of week) in \textbf{TiDE} and \textbf{WaveNet}. Hyperparameters were tuned by random search\cite{bergstra2012random}: for each model, 40 hyperparameter configurations were randomly sampled from a grid, and the optimal configuration was selected as the one with the lowest validation MWQL. Finally, we trained models with the optimal hyperparameter configuration using 5 random seeds, and evaluated them on the test set. Table \ref{tab:hyperparams} lists the names and candidate values of the hyperparameters for each model. Note that the values of \texttt{context\_length} include all candidate values for all prediction lengths, but the candidates may vary across prediction lengths: for each session, we first performed hyperparameter search for forecasting 35 steps. Depending on the validation results, we may adjust the candidates for forecasting 70 and 18 steps. The candidate values of all other hyperparameters were the same across all sessions and prediction lengths.

For \textbf{Chronos}, we used \texttt{chronos-bolt-base} from AutoGluon\cite{agtimeseries}. For fine-tuning, we fixed $\texttt{fine\_tune\_batch\_size} = 128$, $\texttt{fine\_tune\_lr} = 10^{-5}$, and varied $\texttt{fine\_tune\_steps} \in \{100, 200, 400, 600, 800, 1000\}$. The configuration achieving the lowest validation MWQL was then evaluated on the test set. Fine-tuning Chronos yielded identical results across random seeds, likely due to an issue in the AutoGluon implementation.

For \textbf{Moirai}, we used \texttt{moirai-1.0-R-base} from the official implementation of \cite{woo2024unified}. Since \textbf{Moirai} randomly samples prediction and context length during training, for zero-shot evaluation, we performed inference tuning as in \cite{woo2024unified}. Specifically, we varied context length $C \in \{50, 100, 250, 500, 750, 1000, 2000, 3000, 4000, 5000\}$ and patch size $P \in \{8, 16, 32, 64, 128\}$, and evaluated all pairs with the desired prediction horizon on the validation set. The pair with the lowest validation MWQL was then evaluated on the test set. Note that there is not a default patch size for the sampling frequency of widefield data, so we swept through all possible patch sizes of \textbf{Moirai}. 32 and 64 generally work the best. For fine-tuning, we fixed $\texttt{batch\_size} = 64$ (due to memory constraint) and varied $\texttt{learning\_rate} \in \{10^{-3}, 5\times10^{-4}, 10^{-4}, 5\times10^{-5},  10^{-5}, 5\times10^{-6}, 10^{-6}\}$, while keeping other setting as default. The optimal learning rate was chosen as the one with the lowest validation loss when evaluated with the specific prediction length, $\texttt{patch\_sizes} = \left [32, 64 \right ]$, and $\texttt{context\_lengths} = \left [1000, 2000, 3000, 4000, 5000 \right ]$. The best hyperparameter configuration was then fine-tuned with 5 different random seeds. For each seed, we computed the validation MWQL across multiple $(C, P)$ pairs as in the zero-shot setting. The pair with the lowest validation MWQL across 5 seeds was evaluated on the test set. 

\begin{table}[h!]
\centering
\caption{Hyperparameter search space for GluonTS models.}
\label{tab:hyperparams}
\begin{tabular}{lll}
\toprule
Model & Hyperparameter & Values \\
\midrule
\multirow{6}{*}{\textbf{DeepAR}}
 & \texttt{lr}        & \{2.5e-05, 5.0e-05, 7.5e-05, 1.0e-04, 7.5e-04\} \\
 & \texttt{batch\_size}           & \{32, 64, 128\} \\
 & \texttt{hidden\_size}          & \{64, 128, 256\} \\
 & \texttt{num\_layers}     & \{1, 2, 3\} \\
 & \texttt{context\_length}       & \{35, 70, 140, 175, 210, 245\} \\
 & \texttt{lags\_seq}       & \{1,2, \dots, L-1\}, where $L\in \{5, 10, 20, 30, 40\}$ \\
\midrule
\multirow{6}{*}{\textbf{PatchTST}}
 & \texttt{context\_length}       & \{35, 70, 140, 175, 210, 245\} \\
 & \texttt{patch\_len}       & \{24, 32, 40\} \\
 & \texttt{stride}       & $s \times \texttt{patch\_len}$, where $s \in \{0.25, 0.5, 1\}$ \\
 & \texttt{d\_model}       & \{32, 64, 128, 256\} \\
 & \texttt{nhead}       & \{1, 4\} \\
 & \texttt{dim\_feedforward}    & $r \times \texttt{d\_model}$, where $r \in \{1, 2\}$ \\
 & \texttt{activation}       & `relu' \\
 & \texttt{num\_encoder\_layers}       & \{1, 2, 3\} \\
 & \texttt{lr}        & \{0.0001, 0.00025, 0.0005, 0.00075, 0.001\} \\
 & \texttt{batch\_size}           & \{128, 256, 512\} \\
\midrule
\multirow{5}{*}{\textbf{DLinear}} 
 & \texttt{context\_length}        & \{140, 175, 210, 245, 280, 315\} \\
 & \texttt{hidden\_dimension}        & \{16, 32, 64, 128, 256\} \\
 & \texttt{lr}        & \{0.0001, 0.0005\} \\
 & \texttt{kernel\_size}           & \{5\} \\
 & \texttt{batch\_size}           & \{128, 256, 512\} \\
 \midrule
\multirow{5}{*}{\textbf{TiDE}} 
 & \texttt{context\_length}        & \{35, 70, 105, 140, 175, 210\} \\
 & \texttt{feat\_proj\_hidden\_dim}        & 4 \\
 & \texttt{encoder\_hidden\_dim}        & \{64, 128, 256, 512, 1024\} \\
 & \texttt{decoder\_hidden\_dim}        & same as \texttt{encoder\_hidden\_dim} \\
 & \texttt{temporal\_hidden\_dim}        & \{64, 128\} \\
 & \texttt{distr\_hidden\_dim}        & \{4, 8, 16, 32, 64, 128\} \\
 & \texttt{num\_layers\_encoder}        & \{1, 2\} \\
 & \texttt{num\_layers\_decoder}        & \{1, 2, 3\} \\
 & \texttt{decoder\_output\_dim}        & \{4, 8, 16, 32\} \\
 & \texttt{dropout\_rate}        & \{0.3, 0.5\} \\
 & \texttt{num\_feat\_dynamic\_proj}        & 2 \\
 & \texttt{layer\_norm}        & False \\
 & \texttt{lr}        & \{1.e-05, 5.e-05, 1.e-04, 5.e-04\} \\
 & \texttt{batch\_size}           & \{128, 256, 512\} \\
 \midrule
\multirow{5}{*}{\textbf{TFT}} 
 & \texttt{context\_length}        & \{70, 105, 140, 175, 210, 245, 280, 315\} \\
 & \texttt{quantiles}        & [0.05, 0.1, 0.2, 0.3, 0.4, 0.5, 0.6, 0.7, 0.8, 0.9, 0.95] \\
 & \texttt{num\_heads}        & \{1, 2, 4\} \\
 & \texttt{hidden\_dim}        & \{64, 128, 256\} \\
 & \texttt{variable\_dim}        & same as \texttt{hidden\_dim} \\
 & \texttt{dropout\_rate}        & \{0.2, 0.4, 0.6\} \\
 & \texttt{lr}        & \{0.00075, 0.001, 0.0025\} \\
 & \texttt{batch\_size}           & \{128, 256, 512\} \\
  \midrule
\multirow{5}{*}{\textbf{WaveNet}} 
 & \texttt{num\_bins}        & \{1024, 2048\} \\
 & \texttt{num\_residual\_channels}        & \{4, 8, 12, 24, 36, 48\} \\
 & \texttt{num\_skip\_channels}        & \{4, 8, 16, 32, 48, 64\} \\
 & \texttt{dilation\_depth}        & \{2, 4, 8, 10\} \\
 & \texttt{num\_stacks}        & \{1, 2, 3\} \\
 & \texttt{lr}        & \{1.e-05, 5.e-05, 1.e-04, 5.e-04, 1.e-03\} \\
 & \texttt{batch\_size}           & \{32, 64, 128, 256\} \\
\bottomrule
\end{tabular}
\end{table}

\subsection{Supplementary figures}\label{Appendix: suppfigures}

\end{document}